\documentclass{article}
\usepackage{spconf,amsmath,graphicx,url}
\usepackage{xcolor}
\usepackage{subcaption}

\newcommand{\mg}[1]{\textcolor{black}{#1}}

\title{Named Entity Detection and Injection for Direct Speech Translation}
\name{Marco Gaido$^{* \dagger \ddagger}$\thanks{$^{*}$Work done during an internship at Meta AI.}, Yun Tang$^{\star}$, Ilia Kulikov$^{\star}$, Rongqing Huang$^{\star}$, Hongyu Gong$^{\star}$, Hirofumi Inaguma$^{\star}$}
\address{$^{\star}$ Meta AI, USA,
      $^{\dagger}$ Fondazione Bruno Kessler, Italy,
      $^{\ddagger}$ University of Trento, Italy \\
      \texttt{mgaido@fbk.eu},\texttt{\{yuntang,kulikov,rhuangq,hygong,hirofumii\}@meta.com} \\}
\begin{document}
%
\maketitle
\begin{abstract}

In a sentence, certain words are critical
for its semantic.
Among them, named entities (NEs) are 
notoriously challenging for neural models.
Despite their importance, their accurate handling has been 
neglected in speech-to-text (S2T) translation
research, and recent
work has shown that S2T models perform poorly for locations and 
notably person names,
whose spelling is challenging unless known in advance.
In this work, we explore how to leverage
dictionaries of NEs known to likely appear in a given context
to improve S2T model outputs.
Our experiments show that
we can reliably detect NEs likely present in an utterance starting from S2T encoder outputs.
Indeed, we demonstrate that the current detection quality
is
sufficient to improve NE accuracy in the translation
with a 31\% reduction in person name errors.

\end{abstract}
\section{Introduction}
\label{sec:intro}

Translation is the process to convey the same semantic meaning of a source sentence into a target language.
In this process, named entities (NEs) -- which identify real-world people, locations, organizations, etc. --
play a paramount role
and their correct translation is crucial to express the accurate meaning \cite{li-etal-2013-name}.
On the other end, current neural translation systems are known to struggle in presence of rare words \cite{koehn-knowles-2017-six}, as NEs often are.
These motivations drove researchers to study dedicated solutions that exploit additional information available at inference time,
such as bilingual dictionaries \cite{hokamp-liu-2017-lexically,post-etal-2019-exploration,dougal-lonsdale-2020-improving,wang-etal-2022-integrating}.
All these works, however, are targeted for text-to-text (T2T) translation and assume that the dictionary entities present in the source sentence can be easily retrieved with pattern matching.
This assumption does not hold
for the speech-to-text (S2T) translation task, where the source modality is audio.

The S2T task was initially accomplished by a cascade of automatic speech recognition (ASR) and T2T translation systems. However, end-to-end (or direct) S2T solutions have recently progressed up to achieve similar translation quality \cite{bentivogli-etal-2021-cascade}, with the benefits of a simpler architecture and lower latency.
Cascade and direct models have been shown to equally struggle with NEs \cite{gaido-etal-2021-moby}, even more than T2T ones, especially regarding person names \mg{\cite{gaido-etal-2022-talking}} that are particularly hard to recognize from speech. 
Despite this and the importance of NEs, to the best of our knowledge, no work has so far explored how to exploit contextual dictionaries of NEs available at inference time in S2T. In addition,
existing methods designed for T2T are not applicable due to the different input modality.

Motivated by the practical relevance of the problem and the lack of existing solutions, 
we present the first approach to exploit contextual information -- in the form of a bilingual dictionary of NEs -- in direct S2T. Specifically,
our main focus is the detection of the NEs present in an utterance, among those in a given contextual dictionary.
Performing this task
allows us to rely on the existing solutions to inject the correct translations for the NEs.
To showcase that the quality of our NE detector is
sufficient to be useful,
we adopt 
a decoder architecture similar to Contextual Listen Attend and Spell (CLAS) \cite{pundak-etal-2018-clas}
and provide it with the list of translated NEs
considered
present by our detector module.
Experimental results on 3 language pairs (en$\rightarrow$es,fr,it) demonstrate that we can improve NE accuracy by up to 7.1\% over a base S2T model, and reduce the errors on person names by up to 31.3\% over a strong baseline exploiting the same inference-time contextual data.

\begin{figure}[!ht]
    \centering
    \begin{subfigure}[t]{0.23\textwidth}
    \centering
    \includegraphics[width=0.99\textwidth]{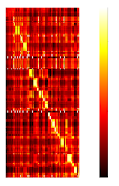}
    \caption{Full sentence.}
    \end{subfigure}
    ~
    \begin{subfigure}[t]{0.22\textwidth}
    \centering
    \includegraphics[width=0.78\textwidth]{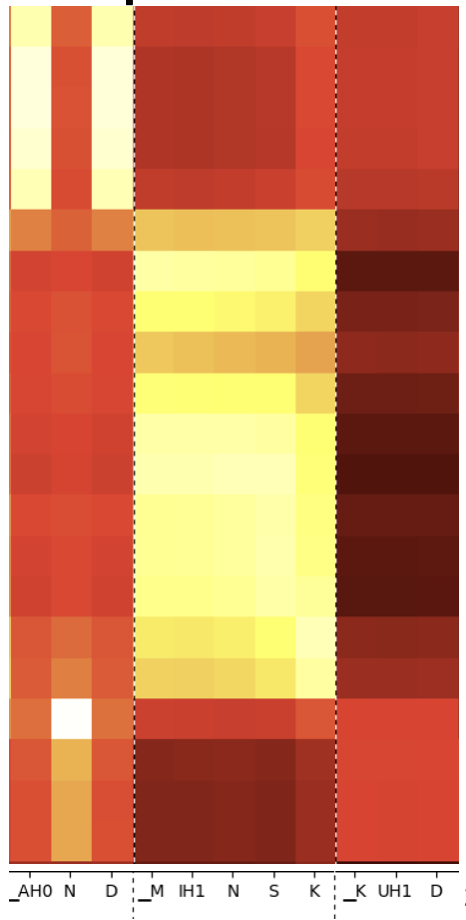}
    \caption{Zoom on \textit{Minsk} phonemes}
    \end{subfigure}
    \caption{Heatmap of cosine similarities (the lighter, the more similar) between the encoder outputs of text and speech of the ST2T model released by \cite{tang-etal-2021-improving}. On the $x$ axis, each item is a phoneme passed to the textual shared encoder; on the $y$ axes there are frames that correspond to the utterance.}
    \label{fig:heatmap}
\end{figure}

\section{Entity Detection for S2T Translation}
\label{sec:entity_detection}

Two operations are necessary to exploit a dictionary of NEs likely to appear in an utterance:
\textit{i)} detect the relevant NEs among those in the dictionary, \textit{ii)} look at the corresponding translations to accurately generate them. Accordingly, we add two
modules 
to the 
S2T model: \textit{i)} a detector identifying
the NEs present in the utterance, and \textit{ii)} a module 
informing the decoder about the forms of the NEs in the target language.

\subsection{Entity Detection}
\label{sec:trained_det}

A recent research direction in S2T consists in training models that jointly perform
S2T and T2T to improve the quality of direct S2T \cite{tang-etal-2021-improving,ye-etal-2022-cross}.
These speech/text-to-text (ST2T) models include auxiliary tasks to force the encoder outputs of different modalities to be close when the text/audio content is the same.
Fig. \ref{fig:heatmap}
confirms
that encoder outputs for text (the text is actually converted into phonemes before being fed to the encoder, as per \cite{tang-etal-2021-improving}) and audio are indeed similar. Specifically, there is a strong similarity between the phonemes that compose a word and the audio frames that correspond to that word.
Based on this, we hypothesize that we can use the encoded representation of the textual NEs in a dictionary/list and the encoded representation of an utterance to determine whether 
each NE has been mentioned or not\mg{.}
To this aim, we train a NE detector module fed with the encoder outputs of different modalities. Such a module, fed with a NE and an utterance, should classify whether the NE is present or not.

At training time, we feed a positive sample (i.e. a piece of text actually present in the utterance) and a negative sample (i.e. a piece of text not present in the utterance) for each audio to train the NE detector.
Positive and negative texts can be sampled \textit{i)} from random words in the transcript of the current utterance and of those of other utterances in the same batch,\footnote{Ensuring they are not present in the examined utterance.} or \textit{ii)} from automatically-detected NEs. The second approach is closer to the real goal, but also limits the amount of training data (ignoring the utterances in the training set that do not contain NEs), and its variety (risking to overfit to the NEs in the training set). 
%
To avoid this, we adopt a mixed approach, where in training samples without NEs the first approach is used, while in training samples with NEs one of the two approaches is randomly selected (assigning 80\% of probability to choosing automatically-recognized NEs).

\begin{figure}[!t]
    \centering
    \includegraphics[width=0.45\textwidth]{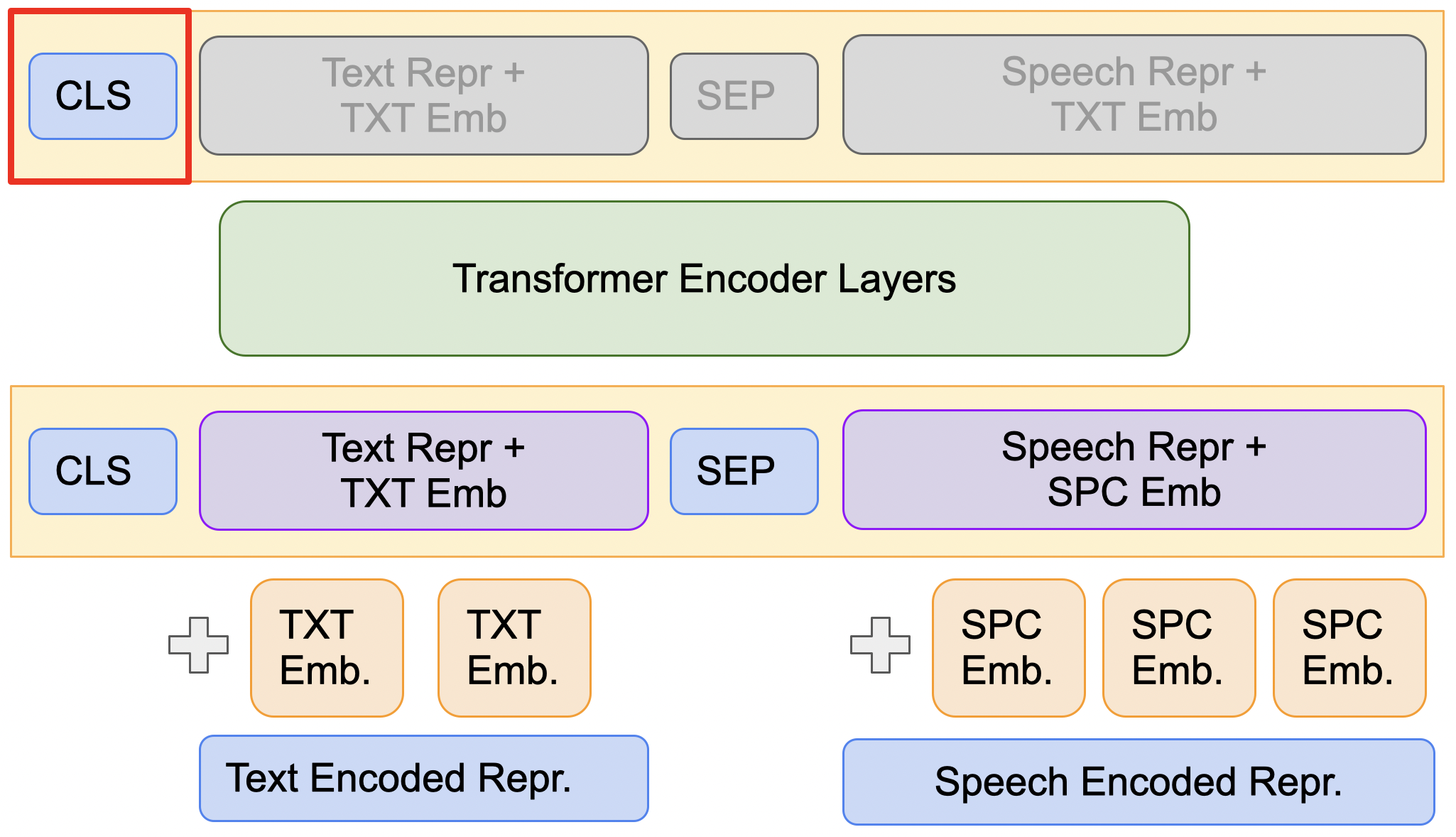}
    \caption{NE Detector architecture.}
    \label{fig:arch}
\end{figure}

Another critical aspect is the design of this detector module.
We first tested the Speech2Slot architecture \cite{Wang-etal-2021-Speech2Slot}, a stack of three layers made of a multi-head attention (MHA) followed by a feed-forward network (FFN) without residual connections. The NE textual encoder output is fed as query to the MHA, while the key and values are built from the speech encoder output.
Unfortunately, training this architecture turned out challenging and the networks failed to converge.

In light of this, we resorted to a stack of three Transformer encoder layers, fed with a concatenation of a \textit{CLS} token, the NE textual encoder output, a \textit{SEP} token, and the utterance S2T encoder output.
From the output of the last layer, we then select only the first vector, corresponding to the \textit{CLS} token, and feed it to a $sigmoid$ ($\sigma$) activation function to get the probability that the NE is present in the utterance.
In addition, we add a trained \textit{TXT} embedding to all the NE textual encoder vectors and a trained \textit{SPC} embedding to all the speech encoder vectors, obtaining the architecture represented in Fig. \ref{fig:arch}.
Finally, as a NE should appear in a contiguous and relatively short speech segment, we force the module to focus on a limited span of speech vectors surrounding the considered one by
an \textit{attention masking} mechanism. Specifically, as the amount of speech that should be considered depends on 
the NE length, we mask all the speech vectors that are further than two\footnote{We also tested 1 and 3, noticing minimal differences and chose 2 due to its lower loss on the dev set.} times the number of phonemes of the NE to detect  with respect to the current speech element. For instance, when trying to detect a NE made of 10 phonemes, each speech vector can attend only to itself, the 20 speech vectors before it, and the 20 after it, in addition to
the textual and the special token vectors. In other words,
each speech vector can attend to the surrounding ones, to all textual vectors, and the \textit{CLS} and \textit{SEP} embeddings.

\subsection{Decoding with Contextual Entities}
\label{sec:clas}

\begin{figure}[!ht]
    \centering
    \includegraphics[width=0.4\textwidth]{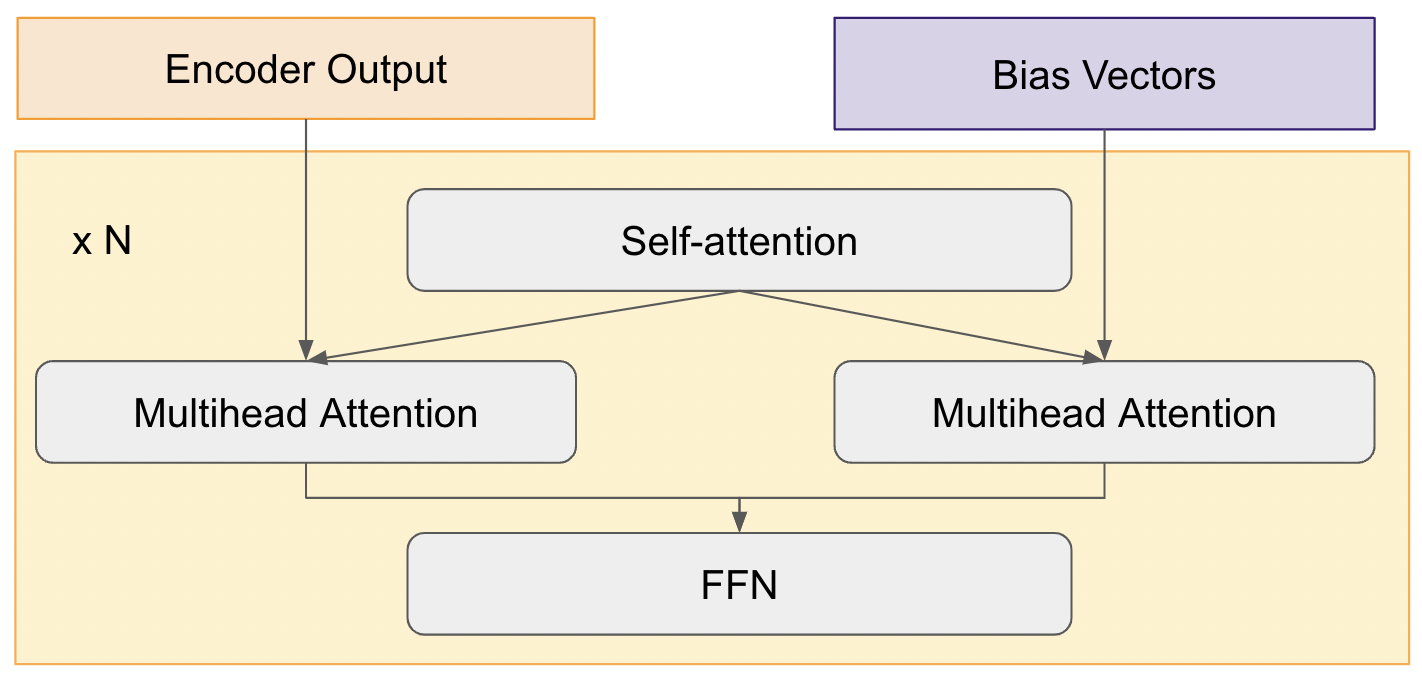}
    \caption{CLAS Transformer decoder layer (\textit{parallel} method).}
    \label{fig:clas}
\end{figure}

As we will see in Sec. \ref{sec:res}, the entity detector achieves high recall, but
false positives are hard to avoid.
Hence, to demonstrate that our entity detector is useful despite a 
low
precision, we inject the selected entities into the model with an approach tolerant to false positives.
%
%
%
We adopt an architecture similar to CLAS \cite{pundak-etal-2018-clas}, where the bias encoder is a trained 3-layer Transformer encoder, and the attention between the decoder and the bias encoder outputs is a MHA implemented following the \textit{parallel} or \textit{sequential} methods by \cite{gaido20_interspeech} (see Fig. \ref{fig:clas}).
Each NE in the list of those considered likely present (\textit{bias-NE})
is encoded with the bias encoder and the encoder outputs are averaged to get a single vector. After repeating this step for all the bias-NEs, the resulting vectors are concatenated together with a \textit{no-bias} learned vector that allows our model to ignore the information from bias vectors.

\section{Experimental Settings}
\label{sec:exp}

Our baselines are a
multilingual direct ST2T and a cascade system (ASR followed by multilingual MT).
ST2T and ASR models are trained according the recipe in \cite{tang-etal-2022-unified}, where the BART text pre-training uses Europarl \cite{koehn-2005-europarl} data,\footnote{We filter from Europarl all the data that belongs to the dates of the talks inside the Europarl-ST test set.} and the joint pre-training uses Libri-Light \cite{librilight} as unsupervised data, and MuST-C \cite{cattoni-2021-mustc} and Europarl-ST \cite{iranzo-2020-europarl-st} as supervised data. For the final fine-tuning, the ASR model is fine-tuned on the en$\rightarrow$es section of MuST-C and Europarl-ST (the largest language direction among the ones we considered), while the ST2T model is fine-tuned on the en$\rightarrow$\{es,fr,it\} S2T directions, together with an auxiliary ASR task that is not used at inference time. Our multilingual MT model is trained on Europarl. All our models also predict NE tags in the output, which has been shown to improve NE handling \cite{xie-2022-e2e-ea-nmt}.

Our ASR and ST2T models directly process raw waveforms using the same hyperparameters of \cite{tang-etal-2022-unified}.
Encoder layers are randomly dropped (LayerDrop) at training time with 0.1 probability \cite{fan2019reducing}. The models have in total $\sim$176M of parameters. The multilingual MT models have 6 encoder and decoder layers with 512 features and 1024 FFN hidden features, for a total of $\sim$102M of parameters.
When training CLAS models, we initialize the weights with those of the pre-trained ST2T model. We 
freeze encoder weights, and we also experimented freezing the decoder parameters from the pre-trained ST2T model: in this case we train only the newly added components and the output projection layer.

The quality of the NE detectors is assessed with the trade-off between recall and number of NEs retrieved. We 
estimate selectivity through the number of NEs retrieved instead of precision, as some NEs may be correctly detected even though they are not annotated in NEuRoparl-ST \cite{gaido-etal-2021-moby},\footnote{\label{foot:nested_ne}For instance, this happens when a NE is part of a bigger one (e.g. the NE \textit{Lisbon} is retrieved in a sentence that contains the NE \textit{Treaty of Lisbon}).} making hard to reliably compute precision.
The output of S2T systems are evaluated with SacreBLEU\footnote{\texttt{case:mixed|eff:no|tok:13a|smooth:exp|v:2.1.0}} \cite{post-2018-call} on Europarl-ST for the translation quality, and with case-sensitive entity accuracy on NEuRoparl-ST for the ability in handling NEs.
Among NEs, we focus on geopolitical entities (GPE), locations (LOC), and person (PER) names, as these three types are the most challenging for S2T systems \cite{gaido-etal-2021-moby}.

\section{Results}
\label{sec:res}

\subsection{NE Detection}
\label{sec:res_detection}

\begin{table}[!bt]
\centering
\small
\begin{tabular}{|l|ccc|r|}
\hline
 & \textbf{GPE} & \textbf{LOC} & \textbf{PER} & \textbf{Retr.} \\ 
\hline
\hline
Cosine Similarities & 68.2\% & 74.3\% & 53.2\% & 138.2 \\
\hline
Base NE detector & 31.4\% & 6.3\% & 28.3\% & 115.5 \\
+ speech masking & 31.9\% & 17.9\% & 29.4\% & 54.3 \\
\hspace{2mm} + layerdrop & 57.2\% & 24.2\% & 38.0\% & 3.9 \\
+ layerdrop & 66.8\% & 33.7\% & 40.2\% & 4.3 \\
\hspace{2mm} + train on NE & 93.9\% & 76.8\% & 79.4\% & 1.8 \\
\hspace{4mm} + modality emb. & 95.2\% & 94.7\% & 78.3\% & 1.8 \\
\hspace{6mm} + attn. masking & 93.5\% & 93.7\% & 90.2\% & 1.6 \\
\hspace{8mm} + max word len 5 & 96.5\% & 93.7\% & 89.1\% & 1.4 \\
\hspace{10mm} + margin ranking & 96.1\% & 91.6\% & 88.0\% & 1.2 \\
\hline
\end{tabular}
\caption{Recalls on GPE, LOC, and PER, and number of NEs retrieved on average (Retr.) for each \mg{utterance.}
}
\label{tab:ne_detect}
\end{table}

Table \ref{tab:ne_detect} reports the retrieval results of the NE detector module described in Sec. \ref{sec:trained_det}, isolating the contribution of 
its components, and compares it with a simple algorithm based on the cosine similarities, 
which is unable to obtain good selectivity.
\mg{For each utterance, the NE detectors are fed with all the distinct GPE, LOC, PER, and organizations (ORG) in the test set for a total of 294 NEs. A NE is considered detected if the NE detector assigns a detection probability higher than 86\%.}
First, we notice that, to achieve meaningful scores, it is essential to introduce LayerDrop when extracting the input features using the shared speech/text encoder of the ST2T model. Otherwise, 
%
%
the results
are close to a random predictor.
Speech masking also 
helps, but 
results harmful when combined with LayerDrop. Moreover, 
feeding automatically-detected NEs at training time with the mixed approach described in Sec. \ref{sec:trained_det},
instead of only using random words, greatly improves both recalls and selectivity.
The addition of trained modality embeddings also proved helpful, especially for LOC and GPE recall. The
attention masking provides significant benefits in terms of PER recall and selectivity, at the cost of a very limited degradation on GPE and LOC recall. Further improvements in selectivity were obtained by picking more than a single random word when training the NE detector (up to 5 consecutive words), and by adding an auxiliary margin ranking loss to the binary cross entropy loss. This final module achieves recalls higher or close to 90\%, 
retrieving 1.2 NEs per utterance on average (the test set contains 0.34 NEs from these 3 categories on average). Excluding
the retrieved NEs present in an utterance but not annotated as such in the test set,\textsuperscript{\ref{foot:nested_ne}} we can compute the precision of this module, which is 55.8\%.
The non-negligible number of false positives is investigated in Sec. \ref{sec:analysis}, and
highlights that the NE detector can be used to create a short-list of NEs likely present in the sentence, rather than enforce the presence of detected NEs, motivating the CLAS solution (Sec. \ref{sec:clas}).

\subsection{S2T Quality and NE Translation}
\label{sec:res_st}

\begin{table}[!tb]
\centering
\small
\begin{tabular}{|l|c|cccc|}
\hline
 & \textbf{BLEU} & \textbf{GPE} & \textbf{LOC} & \textbf{PER} & \textbf{Avg.} \\ 
\hline
Cascade  & 37.6 & 80.0 & 74.2 & 51.2 & 68.5 \\
Base ST2T  & \textbf{38.8} & 82.2 & 78.4 & 49.3 & 70.0 \\
\hline
\hspace{1mm} + CLM ($\lambda$=0.10) & \textbf{38.8} & 83.9 & 76.8 & 50.8 & 70.5 \\
\hspace{1mm} + CLM ($\lambda$=0.15) & 38.0 & 83.6 & 74.9 & 52.7 & 70.4 \\
\hspace{1mm} + CLM ($\lambda$=0.20) & 37.0 & 82.5 & 73.0 & 53.4 & 69.6 \\
\hline
Parallel CLAS & 37.5 & \textbf{84.7} & 78.4 & 66.1 & 76.4 \\
\hspace{1mm} + freeze decoder & 37.0 & 82.8 & 78.7 & 64.6 & 75.4 \\
Sequential CLAS & 35.8 & 84.5 & 78.7 & \textbf{68.0} & \textbf{77.1} \\
\hspace{1mm} + freeze decoder & 36.8 & 82.7 & \textbf{79.9} & \textbf{68.0} & 76.9 \\
\hline
\end{tabular}
\caption{Translation quality (BLEU) and accuracy for GPE, LOC, and PER -- as well as the average over the 3 categories (Avg.) -- of the base direct ST2T, cascade, and the test-entities aware systems (class LM -- CLM -- and
CLAS models). 
The results
are the average over the 3 language
pairs (en$\rightarrow$es,fr,it).
}
\label{tab:clas_res}
\end{table}

Our CLAS method leverages additional data available at inference time. Its comparison with a plain S2T model would hence be unfair, so we introduce a strong baseline that exploits this additional data.
Specifically,
we perform a class language model (LM) rescoring of the S2T model probabilities, using shallow fusion (i.e. adding to the S2T model probabilities the LM probabilities rescored with a weigth $\lambda$) \cite{huang20f_interspeech}.
We train the class LM \cite{huang20f_interspeech} on the test-time NEs, and a generic LM on the target side of the MT training data. At each decoding step, if we are inside NE tags for the current hypothesis, we rescore (shallow fusion) the S2T outputs with the class LM; otherwise, the rescoring is done with the generic LM.

Table \ref{tab:clas_res} compares this strong baseline, the base model, and our CLAS systems fed with the entities selected by our NE detector module. We can see that CLAS systems are the best in NEs accuracy, reducing by up to 31\% the number of errors for person names compared to the best baseline using the same additional information. The improvements for other NEs are lower: we argue that the reason lies in the different representation NEs have in the different languages (source and target) while person names are mostly the same. Despite its better NE handling, CLAS suffers from a 1.3-2.0 BLEU degradation with respect to the baseline and future work should address this weakness.
However, comparing our Parallel CLAS model to the baselines, we notice that the best baseline for person names (CLM $\lambda$=0.20) is significantly inferior on all metrics, including BLEU and person name accuracy. Moreover, BLEU is similar to the cascade solution with significantly higher accuracy on all NE categories.

\section{Analysis}
\label{sec:analysis}

As observed in Sec. \ref{sec:res_detection}, the weakness of the NE detector is the number of wrongly detected NEs (false positives). To better understand why they happen, we conducted a manual analysis of the false positives, assigning each of them to one of the following categories:
\textit{i)} \textbf{similar semantic} (13.7\%), NEs detected in an utterance where there is a NE with a similar meaning (e.g. \textit{Chamber}/\textit{Parliament}) or there is another NE of the same type (e.g. \textit{Pakistan}/\textit{Afghanistan});
\textit{ii)} \textbf{similar phonetic} (14.3\%), NEs detected in sentences where there is a word that is similar or sounds similar (e.g. \textit{President}/\textit{Presidency}); 
\textit{iii)} \textbf{partial match} (34.0\%), NEs detected in utterance where only part of the NE is present (e.g. \textit{Fisheries Committee}/\textit{Budget Committee});
\textit{iv)} \textbf{acronyms} (8.4\%), these NEs are poorly handled because our text-to-phonemes converter does not handle them properly (e.g. \textit{US} is converted as the pronoun \textit{us} and \textit{EU} as the pronoun \textit{you});
\textit{v)} \textbf{different form} (16.5\%), NEs detected where the same NE is mentioned but in a different form (e.g. \textit{government of Malaysia}/\textit{Malaysian Government}), so these are not real errors; 
\textit{vi)} \textbf{uninterpretable} (13.1\%), the human cannot understand the reason of the error.
This inspection shows that future work should focus on training strategies that alleviate the detection errors of similar words and partial matches,
creating systems that are more robust to small yet significant variations between different entities.

\section{Conclusions}
\label{sec:conclusions}

In this work, we explored how to leverage dictionaries of NEs in a specific domain/context
to improve the NE accuracy of S2T systems,
mainly focusing on 
the detection of which NEs of a domain dictionary are present in an utterance.
We proposed an additional module on top of the encoder outputs that can determine whether each NE is present in an utterance,
achieving a high recall for geopolitical entities, locations, and person names.
We demonstrated that the biggest challenge regards increasing the
selectivity of the model, and reported a thorough analysis of the most common false positives with guidelines for future works on the topic. In addition, we 
proposed a method to inject the selected NEs in the decoding phase, 
showing that the proposed detection strategy is already capable of improving NE handling, with average accuracy gains up to 14.4\% on 
GPE, LOC, and PER
over strong baselines leveraging the same inference-time information.


\bibliographystyle{IEEEbib}
\bibliography{refs}

\end{document}